\documentclass[journal]{IEEEtran}
\usepackage{multirow}
\usepackage[table,xcdraw]{xcolor}
\usepackage{graphicx}
\usepackage{amsmath}
\usepackage{amsfonts}
\usepackage{algorithm}
\usepackage{amsthm,amssymb}
\usepackage{mathrsfs}
\usepackage{algorithmic}
\usepackage{makecell}
\usepackage{easyReview}
\usepackage{booktabs}
\setcellgapes{3pt}

\setcellgapes{3pt}

\usepackage[justification=centering]{caption}
\usepackage{color}
\usepackage[subfigure]{tocloft}
\usepackage{subfigure}
\usepackage{tabularx}
\usepackage{amsmath, bm}
\usepackage{hyperref}
\usepackage{xcolor}
\hypersetup{hidelinks,
colorlinks=true,
allcolors=blue,
pdfstartview=Fit,
breaklinks=true}

\usepackage{cite}

\ifCLASSINFOpdf
\else
\fi
\begin{document}
	%
	\title{Large AI Model-Based Semantic Communications}
	
	\author{Feibo Jiang, \textit{Senior Member, IEEE}, Yubo Peng, Li Dong, Kezhi Wang, \textit{Senior Member, IEEE}, Kun Yang, \textit{Fellow, IEEE}, Cunhua Pan, \textit{Senior Member, IEEE}, Xiaohu You, \textit{Fellow, IEEE}
		\thanks{
		Feibo Jiang (jiangfb@hunnu.edu.cn) is with Hunan Provincial Key Laboratory of Intelligent Computing and Language Information Processing, Hunan Normal University, Changsha, China.
		
		Yubo Peng (pengyubo@hunnu.edu.cn) is with School of Information Science and Engineering, Hunan Normal University, Changsha, China.
		
		Li Dong (Dlj2017@hunnu.edu.cn) is with Changsha Social Laboratory of Artificial Intelligence, Hunan University of Technology and Business, Changsha, China.
		
		Kezhi Wang (Kezhi.Wang@brunel.ac.uk) is with the Department of Computer Science, Brunel University London, UK.
		
		Kun Yang (kunyang@essex.ac.uk) is with the School of Computer Science and Electronic Engineering, University of Essex, Colchester, CO4 3SQ, U.K., also with Changchun Institute of Technology.
		
		Cunhua Pan and Xiaohu You (cpan@seu.edu.cn, xhyu@seu.edu.cn) are with the National Mobile Communications Research Laboratory, Southeast University, Nanjing 210096, China.
		}
	}

\markboth{Submitted for Review}%
{Shell \MakeLowercase{\textit{et al.}}: Bare Demo of IEEEtran.cls for IEEE Journals}
%



\maketitle 

\begin{abstract}
Semantic communication (SC) is an emerging intelligent paradigm, offering solutions for various future applications like metaverse, mixed reality, and the Internet of Everything. However, in current SC systems, the construction of the knowledge base (KB) faces several issues, including limited knowledge representation, frequent knowledge updates, and insecure knowledge sharing. Fortunately, the development of the large AI model (LAM) provides new solutions to overcome the above issues. Here, we propose a LAM-based SC framework (LAM-SC) specifically designed for image data, where we first apply the segment anything model (SAM)-based KB (SKB) that can split the original image into different semantic segments by universal semantic knowledge. Then, we present an attention-based semantic integration (ASI) to weigh the semantic segments generated by SKB without human participation and integrate them as the semantic-aware image. Additionally, we propose an adaptive semantic compression (ASC) encoding to remove redundant information in semantic features, thereby reducing communication overhead.
Finally, through simulations, we demonstrate the effectiveness of the LAM-SC framework and the possibility of applying the LAM-based KB in future SC paradigms.

\end{abstract}

\begin{IEEEkeywords}
	Semantic communication; large AI model; knowledge base; SAM; semantic compression.
\end{IEEEkeywords}

%
\IEEEpeerreviewmaketitle

\section{Introduction}
Semantic communication (SC) has recently received much attention as a new intelligent paradigm. It is expected to contribute to various applications such as metaverse, mixed reality (MR), and the Internet of Everything (IoE) \cite{chen2023big}.
Unlike traditional communication methods, which focus on ensuring the accuracy of transmitted bits or symbols, SC prioritizes delivering the intended meaning with minimal data. 
Typically, as shown in Fig. \ref{fig:KBs}, the traditional SC system comprises the following components:
\begin{itemize}
	\item {\it{Semantic encoder}}:
	The semantic encoder extracts semantic information from the original data and encodes these characteristics into semantic features, thus understanding the meanings of data and reducing the scale of the transmitted information from the semantic level.
	
	\item {\it{Channel encoder}}:
	The semantic features could be encoded and modulated by the channel encoder to combat channel impairments and improve robustness, which can ensure data is effectively transmitted on the wireless physical channel. 
	
	\item {\it{Channel decoder}}:
	The channel decoder is used to demodulate and decode the received signal and obtain the transmitted semantic features before the original data are recovered.
	
	\item {\it{Semantic decoder}}:
	The semantic decoder aims to understand the received semantic features, infer the semantic information and recover the original data from the semantic level.
	
	\item {\it{Knowledge base}}:
    The knowledge base (KB) of SC can be seen as a universal knowledge model which can help the semantic encoder and decoder to understand and infer the semantic information effectively.
\end{itemize}

The above components can be implemented by applying deep neural networks (DNNs) which have superior self-learning and feature extraction capabilities. These DNNs can be trained jointly in tandem to maximize expected faithfulness in semantic representation and minimize communications overhead during transmission, and the whole SC system can achieve global optimality. 

Recently, most AI-powered SC system models, including TOSCN \cite{10122232}, DeepSC-ST \cite{10038754}, and DeepJSCC-V \cite{10015684}, centered around designing an efficient communication model. These models heavily rely on the encoder and decoder of SC to extract and interpret semantics. 
The primary model architectures that facilitate this process include encoder-decoder (ED) \cite{luo2022semantic}, information bottleneck (IB) \cite{xie2023robust}, knowledge graph (KG) \cite{li2022cross}, and so on.
Although these methods are capable of extracting semantic information from unstructured data sources, they may not fully exploit the potential benefits of utilizing KB in their approach.

\subsection{Composition of a Universal KB in SC Systems}
KB is essential for SC to distinguish itself from conventional communication systems by its capacity to understand and infer semantic information. We can build a universal KB by learning a large amount of world knowledge, which forms the core of the SC system. The universal KB consists of prior and background knowledge that can be understood and recognized by users.
\begin{enumerate}
	\item {\it{Prior knowledge}}: 
SC defines the structure of semantic representation and the relationships between entities through prior knowledge. For instance, semantic information can be represented in triplet form for image understanding. This means that it is made up of three parts, namely, the objective, attribute, and relationship. The entity typically refers to the nouns in the figure, such as house cat and mouse in Fig. \ref{fig:LAM-SC}. The attribute, on the other hand, is based on the adjectives that describe the entities, for example, ``domestic housecat" or ``clever mouse". Lastly, the relationship refers to the connections between the entities, like ``a housecat catching a mouse" instead of the reverse. In essence, through prior knowledge, machines can effectively communicate with humans through SC based on the same or similar ontology, epistemology and logic. This ensures that the semantic information extracted by the system is fully understood by humans. 
	
\item {\it{Background knowledge}}:
	Semantic information is not just about explicit information, but also involves contexts, implicit meaning and common facts. For example, in Fig. \ref{fig:LAM-SC}, the explicit information is a housecat and a mouse, and the implicit background knowledge is ``Tom and Jerry". Similarly, SC involves the exchange of background knowledge between the sender and receiver such as user identity, interest preferences, and user environments. This facilitates the semantic encoder in extracting the most relevant information that interests both parties and allows the semantic decoder to recover the intended meaning accurately. Essentially, background knowledge acts as a key enabler for the SC model, facilitating accurate semantic extraction, eliminating redundancy, and ensuring a successful semantic alignment between the sender and receiver.
\end{enumerate}

\subsection{Issues about Current KB Schemes in SC Systems}
The current KB schemes in SC are based on mature deep learning technology, which is a data-driven learning process. However, the complicated and time-consuming learning process will result in various issues.
\begin{enumerate}
	\item {\it{Limited knowledge representation}}: 
	Traditional SC systems, normally using DNNs or KGs as the KB, should learn from the environment by supervised learning. However, the layers and parameters of the KB are limited, and the labelled data collected from the environment has a high cost. These KBs with restricted parameters and data prevent them from learning abundant semantic knowledge in large data sets and impair their knowledge representation, as well as hinder their ability to comprehensively capture the underlying meaning of human knowledge. For instance, the words ``apple" in ``Apple Inc." and ``apple soda" may be represented as the same features in the traditional word embedding model.
	\item {\it{Frequent knowledge updates}}:
	Current KB schemes should continuously update their knowledge through training and sharing when the knowledge domain is changed in the environment.
	In real-world scenarios where there is a massive circulation of data, hence frequent updates are required to maintain the performance of the SC system and these updates normally incur huge energy and resource costs, further reducing the efficiency of the KB.
	For example, a translation KB that contains knowledge of specific languages, and needs to update the corresponding knowledge when the dialogue involves new languages, in order to complete the translation tasks.
	\item {\it{Insecure knowledge sharing}}:
	In SC systems, current KBs at the source and destination are different because the environments they perceive are different, which might cause semantic errors. Hence, it is essential to share the KBs between users, and ensure that the sender and receiver are semantically aligned, which in turn necessitates the frequent transmission of knowledge models between different users. These knowledge models might include some highly sensitive human-related information, which could introduce potential privacy and security risks. 
	For instance, a medical KB might contain patient identity, medical history, and other information, where sensitive information might be leaked during KB synchronization.
\end{enumerate}

\subsection{Our Contributions}
Recently, there has been significant progress in the large AI model (LAM), which refers to a type of advanced transformer model with billions of parameters. With the continuous improvement of computing power and the increase in data volume, LAMs have shown great performance in the fields of natural language processing, image and speech recognition, etc. It has many advantages, including \textit{accurate knowledge representation}, \textit{rich prior/background knowledge}, and \textit{low-cost knowledge update}. It presents a new opportunity for addressing the aforementioned issues and enhancing the SC system. In this paper, we present a LAM-based SC (LAM-SC) framework specifically designed for image data transmission. Our contributions can be summarized as follows:

\begin{enumerate}
\item \emph{High-precision semantic segmentation}: We apply a large semantic segmentation model-based KB (SKB), focusing on the SC for image data transmission as an example. SKB leverages the accurate knowledge representation to split a raw or unstructured image into different semantic segments or objectives, each of which can be individually selected and encoded by the sender. This allows the sender to focus on specific semantic objectives that are relevant to their communication requirements. 

\item \emph{Goal-oriented semantic alignment}: We develop an attention-based semantic integration (ASI) mechanism in the SC encoder, which can accurately weigh the semantic importance of the segments generated by the SKB. Then, we integrate the important segments as new semantic-aware data. Therefore, the ASI can realize more precise semantic awareness and thus preserve the critical semantic segments without human intervention.

\item \emph{High-quality semantic compression}: We propose a novel adaptive semantic compression (ASC) encoding scheme in the semantic encoder. The ASC can mask certain parts of transmitted semantic features, and the mask ratio can be adjusted adaptively according to the content of transmitted features. Thus, we can ensure that redundant semantic features are further eliminated, leading to a significant reduction in communication overhead.
\end{enumerate}

The remainder of this paper is organized as follows: We begin by presenting various approaches for implementing LAM-based KBs in SC systems. Following that, we introduce the proposed LAM-SC framework, detailing its key components, such as the SKB, ASI, and ASC schemes. We then conduct simulations to demonstrate the advantages of the LAM-SC framework. Finally, we discuss open issues and conclude this paper.

\section{LAMs-Based KBs in SC systems}
\begin{figure*}[htbp]
	\centering
	\includegraphics[width=18cm]{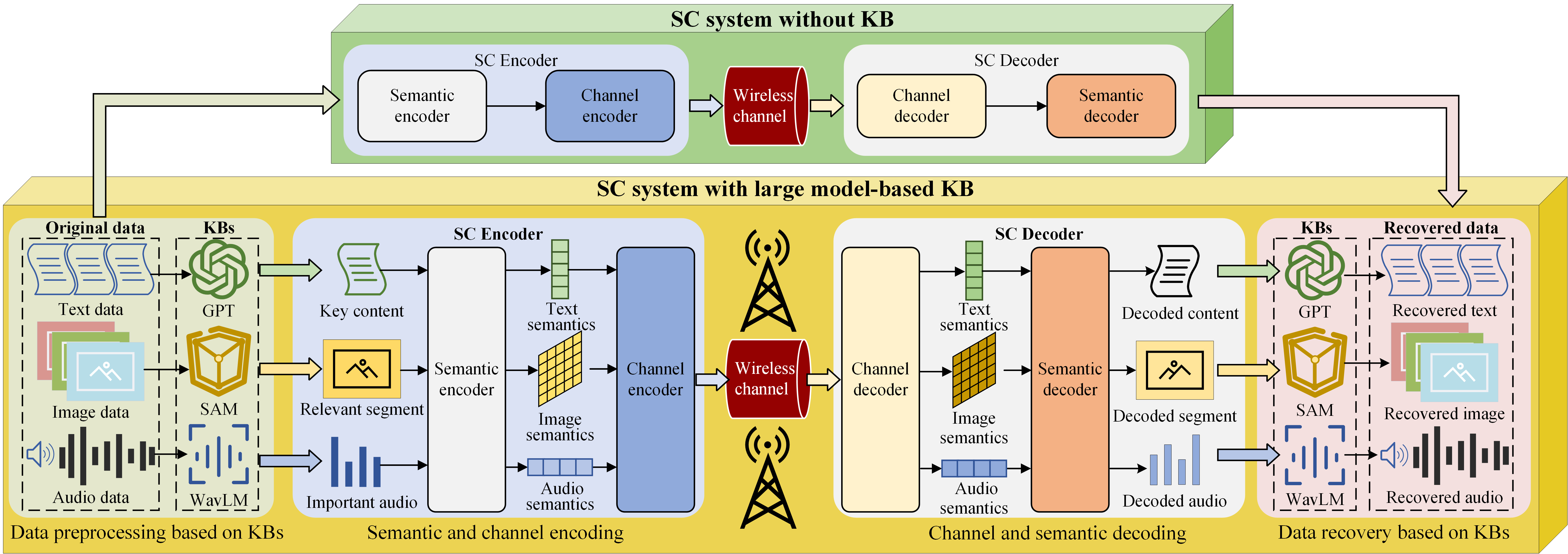}
	\caption{Implementation of LAMs-based KBs in different SC models.}
	\label{fig:KBs}
\end{figure*}

\subsection{Advantages of LAMs-Based KBs}
The LAM refers to the transformer model that has a complicated structure with multi-head attention, enabling it to handle complex AI tasks and generate high-quality outputs. The LAM can be pre-trained on extensive datasets by using self-supervised learning with unlabelled data and then the pre-trained model can be applied to various tasks through prompt learning or fine-tuning. To unlock the potential of LAMs in constructing a more universal KB for the development of SC, we summarize the advantages of introducing LAMs in KBs as follows:
\begin{itemize}
	\item {\it{Accurate knowledge representation}}: Current LAMs, like GPT-4.0, Gemini, and Llama \cite{sung2022vl}, have trillions of parameters, allowing them to learn complex knowledge representations from the transformer model with a multi-head attention mechanism. The multi-head attention mechanism develops a strong understanding of semantics and knowledge structures, hence LAMs can give high-quality semantic representation of input data. For instance, the words ``apple" in ``Apple Inc" and ``apple soda" are represented as different features in LAMs.
	
	\item {\it{Rich prior/background knowledge}}:
	LAMs are pre-trained on extensive datasets such as ImageNet, UCF101, Audioset, and Wikipedia \cite{9770283}, enabling them to learn from the vast amounts of information across various domains. It can store rich prior/background knowledge and have remarkable generalization abilities. It can also achieve high performance on various tasks, even beyond their pre-trained knowledge domains, eliminating the need for frequent updates of KBs.
	
	\item {\it{Low-cost knowledge update}}:
	LAMs typically come with pre-trained weights and can be prompted using just a few examples or fine-tuned with a small amount of labelled data. Techniques like P-Tuning, LoRA, and prompt-tuning could be used for low-cost updates \cite{chowdhury2022novelty}, therefore mitigating concerns of frequent knowledge updates and insecure knowledge sharing.
\end{itemize}

\subsection{Design Suggestions of LAMs in SC Systems}
In this subsection, we suggest several design schemes catering for different types of SC systems (i.e. text, image, audio, etc.), allowing for streamlined integration of LAMs into KB creation, as shown in Fig. \ref{fig:KBs}, 

\subsubsection{GPT-Based KBs}
For the text-based SC system, the KB should be capable of comprehending the text's content and identifying various subjects, their attributes, and relationships. Recently, large language models have emerged, such as ChatGPT \cite{wu2023brief}, which can serve as a semantic knowledge base for text data. ChatGPT is an AI assistant developed by OpenAI based on the GPT-3.5 or 4 model, which can accurately understand the content of the text and provide correct responses to a wide range of questions. By using ChatGPT as the KB for text data, it can extract the key content from the input text based on the users' requirements. In the receiver, the received text data recovered by the SC decoder could be fed to ChatGPT to eliminate semantic noise. Additionally, the received text can be reorganized according to the receiving users' preferences, such as applying different languages.

\subsubsection{SAM-Based KBs}
For the image-based SC system, the KB should be capable of segmenting various objectives in an image and recognizing their respective categories and interrelationships. One promising AI model that can be applied here is the Segment Anything Model (SAM) which is introduced by Meta AI \cite{kirillov2023segment}. SAM is a groundbreaking segmentation system that can generalize zero-shot results to unfamiliar images and objectives without any additional training.
Therefore, SAM may be considered as the perfect KB for images in SC. For real systems, the sender can use SAM to segment the input image and select the important and meaningful segments for the SC encoder. On the receiver side, 
the SC decoder generates the restored image data, subsequently mitigating semantic noise or interference using SAM. This enables efficient identification and extraction of the pertinent segments.
\subsubsection{WavLM-Based KBs}
To enable the SC system for audio, the KB should be capable of performing a variety of tasks, including automatic speech recognition, speaker identification, and speech separation. This ensures that the raw audio data can be analyzed and semantic information can be extracted effectively.
WavLM \cite{chen2022wavlm}, as a large-scale audio model proposed by MSRA, could be one potential solution. Trained on 94,000 hours of unsupervised English data, WavLM is highly effective across a range of speech recognition tasks and non-content recognition speech tasks.
By using WavLM as the KB, the sender can first separate and recognize the audio data from different speakers, discarding unimportant information such as background noise. The remaining audio data is then integrated and encoded by the SC encoder. On the receiver side, the SC decoder can recover the audio, followed by speech de-noising and recognition by WavLM based on the user's requirements.

Among the above-mentioned LAM schemes, the WavLM-based SC is well-suited for real-time interactions and instant communication, enabling quick and efficient information exchange. In contrast, the GPT-based SC excels at conveying thoughts and ideas clearly through textual information, making it easy to store, retrieve, and analyze text data. The SAM-based SC system focuses on transmitting visual information via images, capturing intricate details, spatial organization, and colour, as well as accurately representing expressions, emotions, and non-verbal cues for a more intuitive communication experience. 
At present, there is relatively little research on image-based SC, which motivates this work.

\section{Architecture of LAM-SC Framework}
\begin{figure*}[htbp]
	\centering
	\includegraphics[width=18cm]{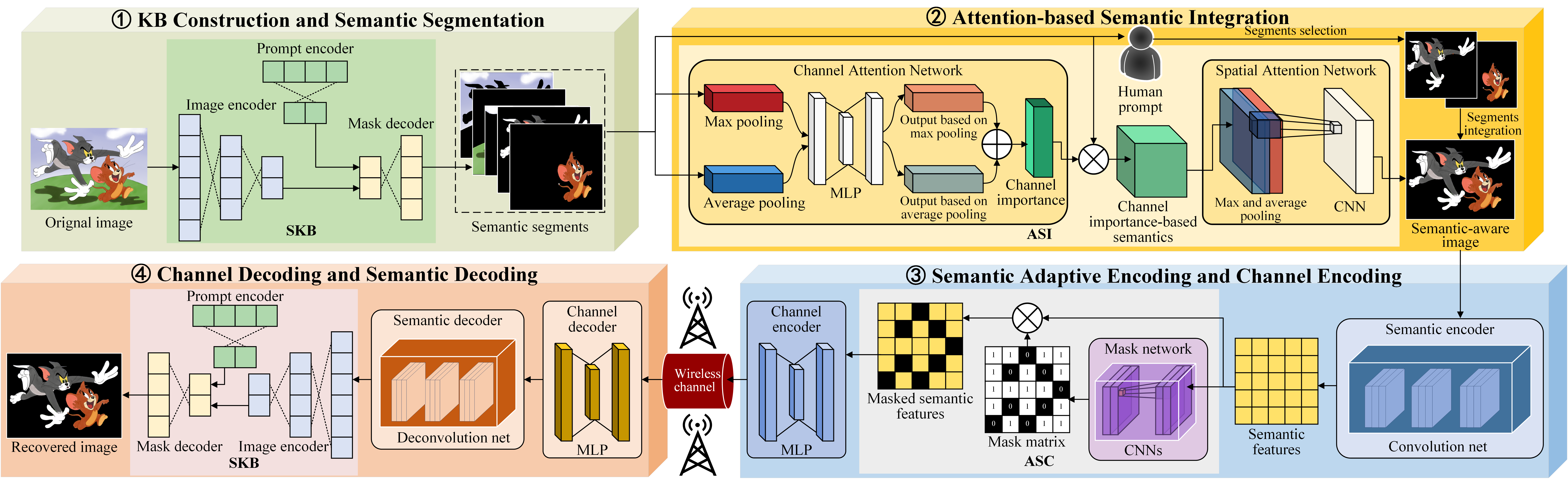}
	\caption{The illustration of the proposed LAM-SC framework.}
	\label{fig:LAM-SC}
\end{figure*}
Introducing the LAMs into SC systems is a promising solution to realize more precise semantic awareness and universal KB in the image-based SC. In this section, we present the LAM-SC framework based on image data that incorporates the SAM model in SC systems. The workflow of the LAM-SC framework is shown in Fig. \ref{fig:LAM-SC}. 

\subsection{Introduction to LAM-SC Framework}
\subsubsection{KB Construction and Semantic Segmentation}
To achieve semantic segmentation of arbitrary original images without the need for KB training, the SKB can be applied to enable comprehensive recognition and segmentation of all semantic objectives in an input image. This process involves analyzing the visual information conveyed by the image to identify each individual objective. As a result, multiple segments may be generated, with each containing only one semantic objective. 

\subsubsection{Attention-Based Semantic Integration}
The ASI can be used to simulate human perception to select the semantic segments that are most worthy of concern by channel attention and spatial attention.
Additionally, we give a human prompt way to obtain interested semantic segments directly. Here, the human prompt refers to the selection result based on human intention.
As a result, the selected segments can then be merged as a new semantic-aware image. 

\subsubsection{Semantic Adaptive Encoding and Channel Encoding}
The semantic-aware image is encoded into semantic features by the semantic encoder. Here, the semantic encoder is built based on convolutional neural networks (CNNs) that have excellent extraction capabilities of image features. 
Moreover, the ASC can be applied to adaptively mask the unimportant features of the semantic information according to its content. Then, the channel encoder that builds based on the multilayer perceptron (MLP) can be used to perform signal encoding and modulation for the physical channel.

\subsubsection{Channel Decoding and Semantic Decoding}
This module performs signal demodulation and decoding functions to obtain the semantic features when the transmitted signals reach the receiver through the wireless physical channel.
Specifically, the channel decoder adopts the MLP architecture to obtain the semantic features. The semantic decoder which consists of deconvolution layers decodes the semantic features and then recovers the image data. Next, the SKB can be employed again on the recovered image to identify objectives accurately.
 

\subsection{SKB}
To achieve image semantic segmentation for input images without specific training, we employ SAM as the KB in our proposed LAM-SC framework, namely SKB, as depicted in Fig. \ref{fig:LAM-SC}, which handles the image semantic segmentation process. SAM is a revolutionary segmentation system that is trained on the largest and most comprehensive dataset. Segment Anything 1-Billion (SA-1B) contains over 1 billion masks across 11 million licensed and privacy-conscious images \cite{kirillov2023segment}. This breakthrough system can successfully generalize zero-shot segmentation for previously unseen images or objectives without requiring additional knowledge and training. 
 
SKB utilizes an efficient transformer-based architecture, designed for both natural language processing and image recognition tasks \cite{kirillov2023segment}. The system comprises a visual transformer-based image encoder for feature extraction, a prompt encoder for user engagement, and a mask decoder for segmentation and confidence score generation.
Here, we apply SAM to automatically achieve objective separation, producing multiple semantic segments for further analysis and processing. 

\subsection{ASI}
The attention mechanism mimics human vision, focusing on crucial details while ignoring irrelevant content. The ASI introduces an attention mechanism to identify and weigh significant objectives in images, which consists of two parts:
\subsubsection{Channel Attention Network}
Using the channel attention network, one can obtain the important parts of each semantic segment. Each segment is treated as a channel, with max and average pooling operations performed. The results are then input into an MLP network for assessing the channel significance. The MLP outputs are combined to determine the importance of each channel, which is then multiplied by the semantic segments to obtain the channel importance-based semantics. 

\subsubsection{Spatial Attention Network}
The above-mentioned channel importance-based semantics can only represent the importance of the semantic segments, however it may ignore the importance from the spatial perspective. To address this, we further use the spatial attention network to process channel importance-based semantics. Specifically, we separately perform max and average pooling on channel importance-based semantics and concatenate results along the channel dimension. 
Next, we integrate the concatenate results and input them into the linear layer-based output head. Then we can obtain the selection results of the semantic segments. Lastly, we integrate the selected segments to obtain semantic-aware images.

In summary, the ASI is capable of intuitively recognizing and retaining essential objects in original images that are typically of greater interest to humans, even without any human involvement.

\subsection{ASC}
ASC method is applied to adaptively mask the transmitted semantic features from the semantic level, effectively reducing the redundant data and significantly decreasing communication overhead. As illustrated in Fig. \ref{fig:LAM-SC}, we utilize a learnable mask network to generate the mask matrix, thereby eliminating unimportant data from the encoded semantic features. During transmission, the encoded semantic features are fed into the mask network, which outputs a corresponding mask matrix with values of either 0 or 1. The semantic features are then multiplied by the mask matrix, causing a portion of the unimportant features to be set to 0 and then obtaining the masked semantic features.

To sum up, by applying the ASC to the semantic transmission process, essential semantic features can be maintained while redundant semantic features are excluded, leading to a substantial reduction in communication overhead.

\section{Training of LAM-SC Framework}
\begin{figure*}[htbp]
	\centering
	\includegraphics[width=18cm]{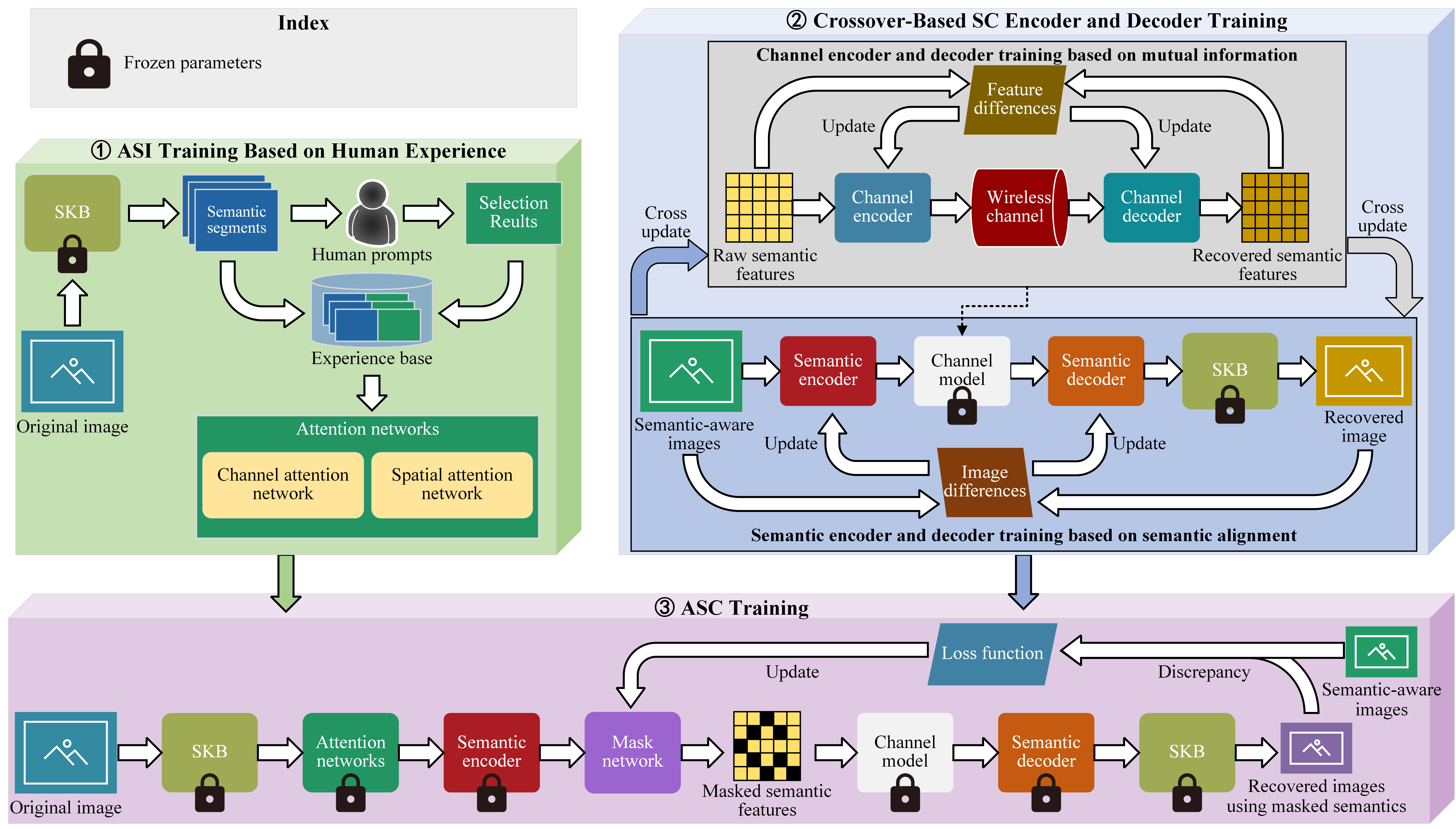}
	\caption{The training process of the proposed LAM-SC framework.}
	\label{fig:LAM-SC_train}
\end{figure*}
In this section, we show how to train the proposed LAM-SC framework, as illustrated in Fig. \ref{fig:LAM-SC_train}, with SAM in SKB as a pre-trained LAM.
\subsubsection{ASI Training Based on Human Experience}
As mentioned before, the aim of the proposed ASI is to mimic human perception in identifying interested objectives in original images, and then producing semantic-aware images that correspond to human preferences. To achieve this, we record human-interested semantics as experiences, which form the foundational training for the attention networks, encompassing both channel and spatial attention networks. In this experience base, semantic segments can serve as input samples for attention networks, while the selection results based on human prompts can be seen as associated labels. By using supervised learning on the experience database, attention networks can effectively adapt to human behaviour and make decisions that closely resemble human perception. 

\subsubsection{Crossover-Based SC Encoder and Decoder Training}
The SC encoder consists of semantic and channel encoders, while the SC decoder comprises channel and semantic decoders. 
Firstly, we use the difference between the raw and the recovered semantic features as the loss function to guide the training of the channel encoder and decoder. 
Then, we use the difference between the original and recovered images as the loss function to guide the semantic encoder and decoder learning training.
We implement a crossed-training strategy involving both the channel encoder/decoder and semantic encoder/decoder models. To be more specific, we first train the channel encoder/decoder,  then freeze its parameters, and next train the semantic encoder/decoder. Next, we freeze the semantic model parameters and train the channel model again. This process can be repeated until the entire SC model achieves convergence.

\subsubsection{ASC Training}
For generating a mask array that accurately reflects the importance of semantic features, we propose a joint training approach for the mask network and the SC model (i.e., SC encoder/decoder), where the parameters of the SC model and the attention networks are frozen. The training process includes the following steps:
1) Transmit the masked semantic features compressed by the mask network; 
2) Recover the image using the received masked semantic features;
3) Calculate the difference between the reconstructed image and the transmitted original image (i.e., semantic-aware image) based on the loss function and guide the training of mask network, enabling it to learn how to generate the optimal mask matrix that minimizes this difference.

\section{Simulation Results}
We use the VOC2012 and COCO2017 \cite{xu2021training} datasets to evaluate the effectiveness of the proposed framework.
For the LAM-SC architecture, the following components are included:
\begin{enumerate}
\item \emph{Attention network}: 
The channel network is composed of two blocks: the first block incorporates a max pooling layer followed by two convolutional layers, while the second block consists of an average pooling layer and two convolutional layers. The spatial network is structured with an average pooling layer, a max pooling layer, and a convolutional layer.
	\item \emph{Semantic encoder}: It comprises two blocks, each with a convolutional layer and a pooling layer.
	\item \emph{Mask network}: It consists of two convolutional layers, with each layer being followed by a Rectified Linear Unit (ReLU) activation function.
	\item \emph{Channel encoder/decoder}: 
	The design of the channel model, encompassing channel encoding and decoding as well as wireless channel configuration, adopts similar settings to those presented in \cite{xie2021deep}.
	\item \emph{Semantic decoder}: It consists of two blocks, each with a deconvolution layer and an upsampling layer.
\end{enumerate}

The compared methods include the traditional SC, LAM-SC, and SC with ASC. The traditional SC and SC with ASC schemes do not utilize the SKB and ASI for processing raw images before transmission, and both of them transmit the original images. On the other hand, the LAM-SC aims to transmit semantic-aware images.
We evaluate performance using two key metrics: system loss and structural similarity (SSIM). 

\begin{figure}[htbp]
	\centering
	\includegraphics[width=8.5cm]{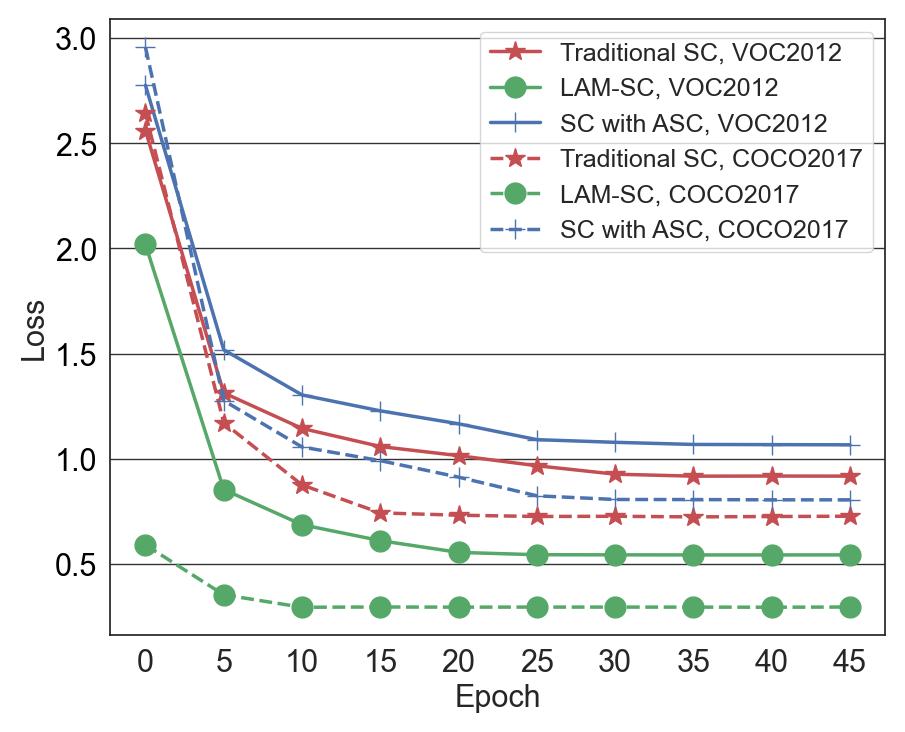}
	\caption{The system loss results of different schemes.}
	\label{fig:exp1}
\end{figure}
\begin{figure}[htbp]
	\centering
	\includegraphics[width=8.5cm]{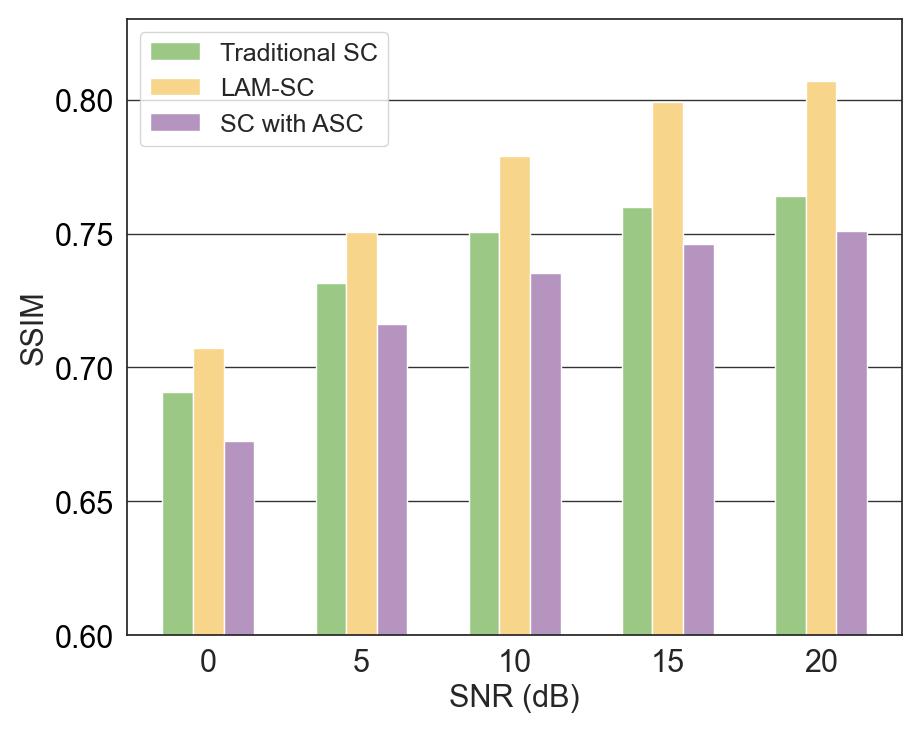}
	\caption{The SSIM results of different schemes on VOC2012 dataset.}
	\label{fig:exp2}
\end{figure}
\begin{figure}[htbp]
	\centering
	\includegraphics[width=8.5cm]{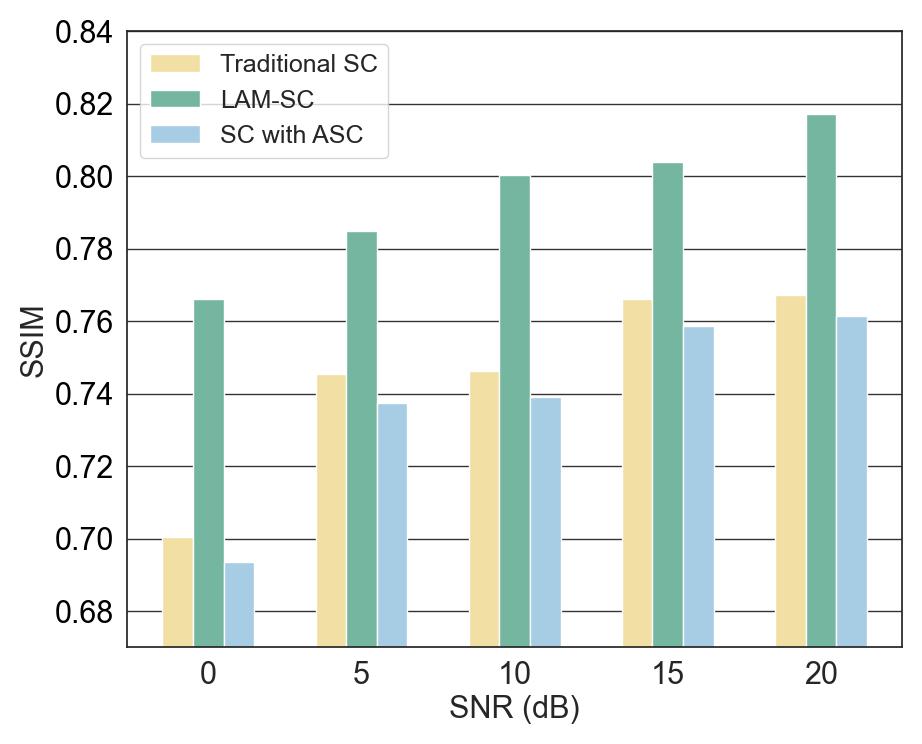}
	\caption{The SSIM results of different schemes on COCO2017 dataset.}
	\label{fig:exp3}
\end{figure}

Fig. \ref{fig:exp1} illustrates the system loss of different schemes when SNR = 5 dB. We can see the LAM-SC obtains the lowest loss result while the traditional SC and SC with ASC schemes have worse performances. 
Fig. \ref{fig:exp2} and Fig. \ref{fig:exp3} describe the SSIM results with different SNRs on VOC2012 and COCO2017 datasets, where one can see that LAM-SC achieves the best performance.

Additionally, in our simulations, the original image in Fig. \ref{fig:LAM-SC} requires 49,152 bits for transmitting, whereas the semantic features transmitted in the traditional SC are 21,632 bits, and the semantic features transmitted in LAM-SC only need 8,960 bits due to the use of ASC. Then, to make a fair comparison, the experiment is conducted on semantic-aware images for both LAM-SC and SC, achieving similar SSIM values. However, the amount of data transmitted by LAM-SC is only 55\% of that of SC.

 
\section{Open Issues and Future Directions}

\subsection{Latency in Real-Time Applications}
LAMs with millions or billions of parameters require substantial runtime, resulting in significant latency during training, updating, and decision-making processes. Furthermore, bandwidth limitations in communication systems could lead to bottlenecks when transferring considerable amounts of data from LAMs in SC systems. Reducing latency is crucial for real-time applications such as metaverse and MR, where immediate responses are essential.  
In the future, some fast inference techniques, such as speculative sampling and multi-query attention, may be applied to the KB to reduce the computation latency.

\subsection{Energy Consumption for Mobile Devices}
The implementation of LAMs in SC systems requires a significant level of energy consumption compared with traditional methods. It raises environmental concerns and presents accessibility challenges for mobile and edge devices. As a result, striking a balance between computational demands and energy constraints is of critical importance. 
In the future, further exploration of compression techniques such as sparse attention, quantization, and distillation may be necessary, as they could provide potential means for reducing the model size and energy consumption for mobile/edge devices.

\subsection{Explainability and Transparency}
Interpreting the outputs of the KB and semantic encoder made by LAMs could be difficult. The LAMs often lack interpretability, making it challenging to understand the semantic analysis process. This can pose difficulties in identifying potential biases or errors in SC systems. 
Adding explainable AI technologies, such as fuzzy rules and visual heat maps into the design of SC systems could be a possible research direction.

\subsection{Privacy and Security}
LAMs can capture sensitive information during training or infer sensitive details from the data they process. Integrating these models into communication systems raises concerns about privacy and security. Ethical considerations regarding issues like consent and responsible use become critical in SC. 
One future research direction could be applying federal learning to perform the fine-tuning of LAMs and SC systems, as well as using differential privacy, homomorphic encryption, and other technologies during the parameter or information exchange process.

\section{Conclusion}
In this paper, we first introduce the importance and composition of KBs, and then discuss the issues about current KB schemes in SC systems. To address these issues, we recommend applying LAMs in building KBs, and we then explore several LAM-based schemes to realize KBs in different SC systems. Next, we propose a LAM-SC framework focusing on image data transmission, where the SAM is applied as the KB for high-quality semantic segmentation, and ASI is presented to integrate segment semantics as a new semantic-aware source. Additionally, ASC is proposed to reduce communication overhead in semantics transmission. Finally, we conduct simulations to demonstrate the effectiveness of the proposed LAM-SC framework.


\bibliographystyle{ieeetran}
\bibliography{bare_jrnl_bobo}
\section*{Biographies}
\textbf{Feibo Jiang} received his B.S. and M.S. degrees in School of Physics and Electronics from Hunan Normal University, China, in 2004 and 2007, respectively. He received his Ph.D. degree in School of Geosciences and Info-physics from the Central South University, China, in 2014. He is currently an associate professor at the Hunan Provincial Key Laboratory of Intelligent Computing and Language Information Processing, Hunan Normal University, China. His research interests include artificial intelligence, fuzzy computation, Internet of Things, and mobile edge computing.

\textbf{Yubo Peng} received the B.S. degree from Hunan Normal University, Changsha, China, in 2019, where he is currently pursuing the master’s degree with the College of Information Science and Engineering. His main research interests include federated learning and semantic communication.

\textbf{Li Dong} received the B.S. and M.S. degrees in School of Physics and Electronics from Hunan Normal University, China, in 2004 and 2007, respectively. She received her Ph.D. degree in School of Geosciences and Info-physics from the Central South University, China, in 2018. She is currently an associate professor at Hunan University of Technology and Business, China. Her research interests include machine learning, Internet of Things, and mobile edge computing.

\textbf{Kezhi Wang} received a PhD degree in Engineering from the University of Warwick, U.K. Currently, he is a Senior Lecturer at the Department of Computer Science, Brunel University London, U.K. His research interests include wireless communications, mobile edge computing, and machine learning.

\textbf{Kun Yang} received his PhD from the Department of Electronic Electrical Engineering of University College London (UCL), UK. He is currently a Chair Professor in the School of Computer Science Electronic Engineering, University of Essex, leading the Network Convergence Laboratory (NCL), UK. He is also an affiliated professor at UESTC, China. Before joining in the University of Essex at 2003, he worked at UCL on several European Union (EU) research projects for several years. His main research interests include wireless networks and communications, IoT networking, data and energy integrated networks and mobile computing. He manages research projects funded by various sources such as UK EPSRC, EU FP7/H2020 and industries. He has published 400+ papers and filed 30 patents. He serves on the editorial boards of both IEEE (e.g., IEEE TNSE, IEEE ComMag, IEEE WCL) and non-IEEE journals (e.g., Deputy EiC of IET Smart Cities). He was an IEEE ComSoc Distinguished Lecturer (2020-2021). He is a Member of Academia Europaea (MAE), a Fellow of IEEE, a Fellow of IET and a Distinguished Member of ACM. 

\textbf{Cunhua Pan} received the B.S. and Ph.D. degrees from the School of Information Science and Engineering, Southeast University, Nanjing, China, in 2010 and 2015, respectively. He was a Research Associate with the University of Kent, Canterbury, U.K., from 2015 to 2016. He held a postdoctoral position with the Queen Mary University of London, London, U.K., from 2016 and 2019, and was a Lecturer from 2019 to 2021. He has been a Full Professor with Southeast University since 2021. He has published over 120 IEEE journal papers. His research interests mainly include reconfigurable intelligent surfaces (RIS), intelligent reflection surface, ultrareliable low-latency communication, machine learning, UAV, Internet of Things, and mobile-edge computing.

\textbf{Xiaohu You} received the M.S. and Ph.D. degrees in electrical engineering from Southeast University, Nanjing, China, in 1985 and 1988, respectively. Since 1990, he has been with the National Mobile Communications Research Laboratory, Southeast University, where he is currently the Director and a Professor. From 1999 to 2002, he was a Principal Expert of the C3G Project, responsible for organizing China 3G Mobile Communications Research and Development Activities. From 2001 to 2006, he was a Principal Expert of the China National 863 Beyond 3G FuTURE Project. Since 2013, he has been a Principal Investigator of the China National 863 5G Project. He has contributed over 200 IEEE journal articles and two books in the areas of adaptive signal processing and neural networks, and their applications to communication systems. His research interests include mobile communication systems, and signal processing and its applications.
Dr. You was selected as an IEEE Fellow for his contributions to the development of mobile communications in China in 2011. He was a recipient of the National 1st Class Invention Prize in 2011. 

\newpage
\end{document}